\documentclass[10pt,twocolumn,letterpaper]{article}
\PassOptionsToPackage{table}{xcolor}
\usepackage{cvpr}
\usepackage{marvosym}
\usepackage[T1]{fontenc}
\usepackage{amsmath,amssymb}
\usepackage{graphicx}
\usepackage{booktabs,multirow,array}
\usepackage[table]{xcolor}
\usepackage{xspace}
\usepackage{microtype}
\definecolor{cvprblue}{RGB}{44,102,156}
\definecolor{direteal}{RGB}{30,115,125}
\definecolor{direlight}{RGB}{233,244,244}
\definecolor{rimehighlight}{RGB}{233,231,250}
\definecolor{teacherorange}{RGB}{191,111,42}
\definecolor{draftgray}{RGB}{103,112,120}
\newcommand{\method}{DiLE\xspace}
\newcommand{\oursrow}{\rowcolor{direlight}}

\usepackage[pagebackref,breaklinks,colorlinks,allcolors=cvprblue]{hyperref}
\title{DiLE: Discrimination-Aware Latent-Space Learning \\ for Universal Multimodal Embeddings}

\author{
Ben Kang$^{1,*,\ddagger}$, Junda Wang$^{2,*}$, Yuxuan Yuan$^3$, Hang Li$^3$, Yuhan Liu$^3$, Xin Chen$^4$, Wanting Geng$^1$,\\
Tong Lin$^2$, Pei Fu$^{3,\dagger}$, Dong Wang$^{1,\dagger}$, Jian Luan$^3$, Huchuan Lu$^1$\\
$^1$Dalian University of Technology,
$^2$Peking University,
$^3$Xiaomi Inc,
$^4$City University of Hong Kong\\
{\tt\small kangben@mail.dlut.edu.cn, wdice@dlut.edu.cn, fupei1@xiaomi.com}
}

\begin{document}
\maketitle
\begin{abstract}
Multimodal large language models have greatly improved input understanding in universal multimodal embedding models. However, this understanding does not ensure that their embeddings preserve the evidence needed to distinguish relevant items from confusable alternatives. To preserve essential discriminative evidence in embeddings, this paper proposes DiLE, a discrimination-aware latent-space learning framework that combines task guidance with on-policy retrieval feedback.
Specifically, a few learnable queries are used to encode task-relevant information in a compact latent space. Reasoning instructions and auxiliary answer-generation supervision shape these representations to support both task understanding and retrieval discrimination. To address task-specific discriminative needs, we design shared–private expert adaptation that integrates cross-task knowledge sharing with task-specific representation learning. Building on this, the on-policy hard-negative distillation is introduced: the student retrieves confusable candidates, and a reverse KL divergence objective aligns its embedding-based similarity distribution with the teacher’s fine-grained relevance distribution over these candidates. Candidates are periodically refreshed using updated student snapshots, tying teacher supervision to the student’s evolving retrieval behavior and confusions.
Across four model configurations that span two backbone generations, DiLE achieves state-of-the-art performance on MMEB-v2, UVRB, and MRMR, demonstrating its effectiveness across modalities and tasks.
\end{abstract}
\vspace{-6mm}
\renewcommand{\thefootnote}{}
\footnote{
\hspace*{-1.8em}$^*$Equal contribution.\\
\textsuperscript{$\dagger$}Corresponding author.\\
\textsuperscript{$\ddagger$}Work done during internships at Xiaomi Inc.
}
\begin{figure}[t]
\centering
\includegraphics[width=\linewidth]{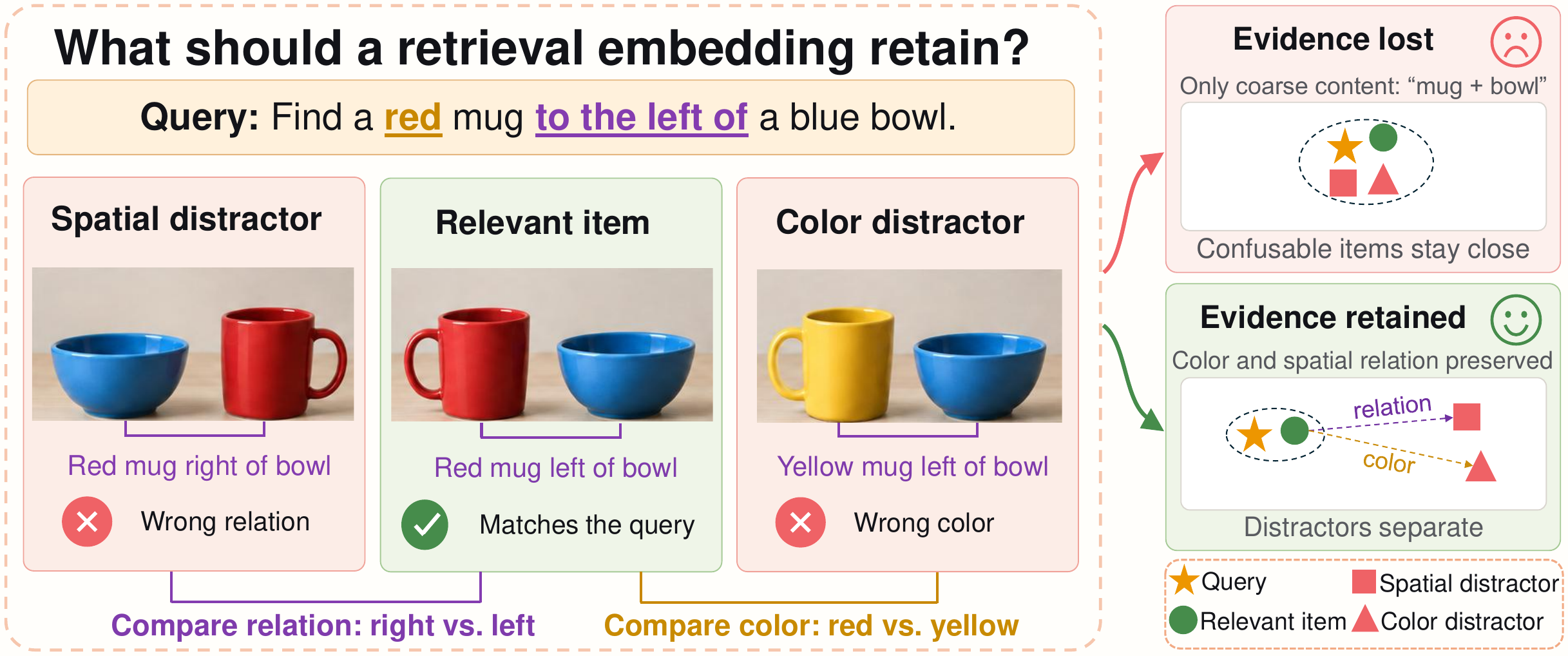}\par
{\small (a) Preserving evidence for retrieval discrimination.\par}
\includegraphics[width=0.85\linewidth]{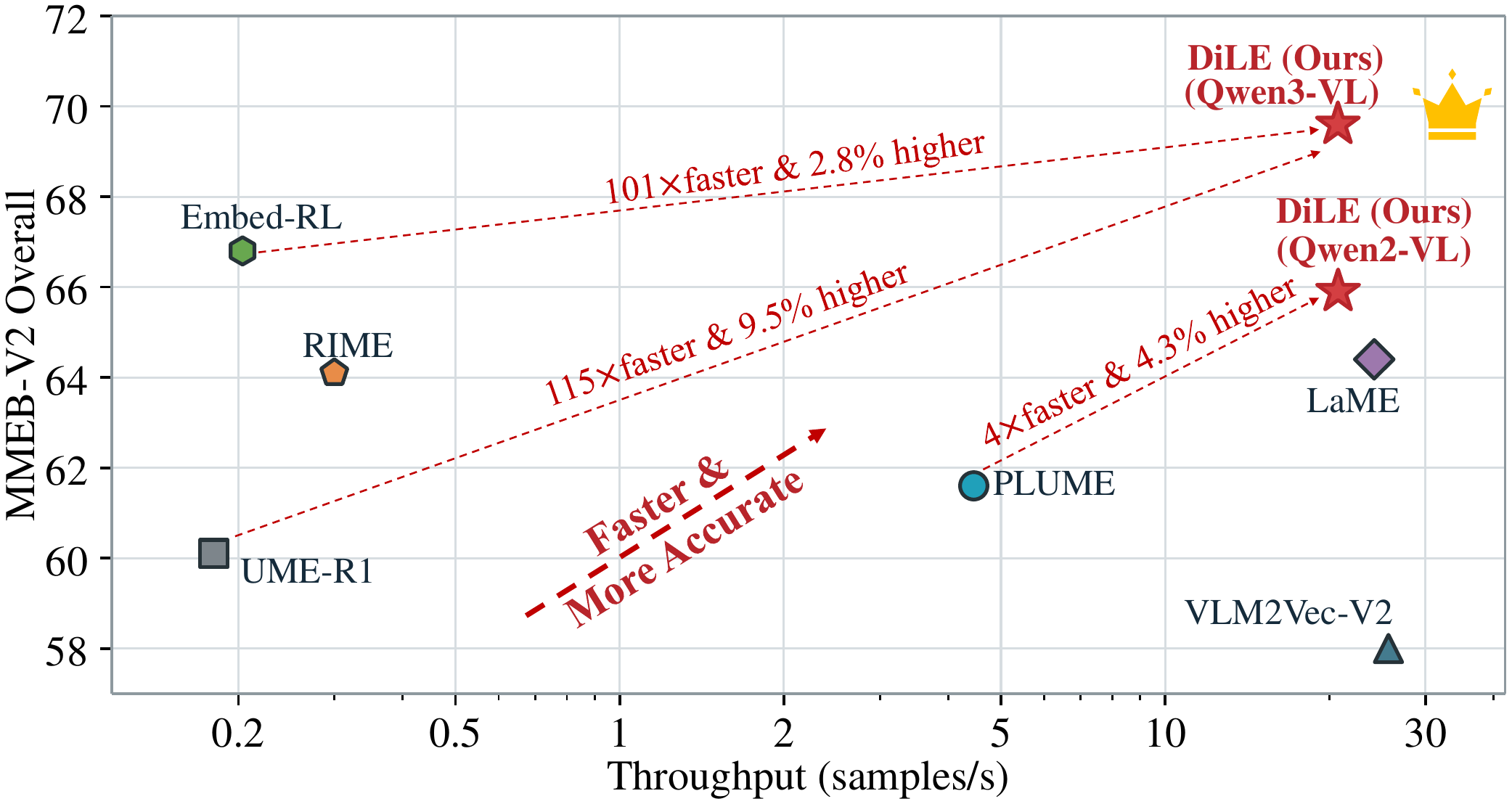}\par
{\small (b) Comparison of throughput and retrieval performance.\par}
\caption{Discrimination-aware UME. (a) Color and spatial relations distinguish the relevant item from confusable alternatives. (b) MMEB-V2 overall performance versus encoding throughput for DiLE and existing embedding models. Throughput was measured on an NVIDIA H100 80GB HBM3 GPU.}
\label{fig:teaser}
\vspace{-4mm}
\end{figure}
\section{Introduction}
\label{sec:intro}
Embedding models enable efficient retrieval by mapping retrieval queries and candidates to vectors that can be compared directly. Multimodal Large Language Models (MLLMs) have advanced Universal Multimodal Embeddings (UME), enabling one model to handle diverse modalities and tasks~\cite{moca,mmE5,zhang2024gme,elva}. Recent methods further use generated reasoning~\cite{lan2025ume} or latent representations~\cite{plume} to improve retrieval. These advances strengthen the model's understanding of multimodal inputs and task instructions. For retrieval, however, this understanding must also be reflected in the distances between candidate embeddings. As shown in Figure~\ref{fig:teaser}(a), candidates may contain similar objects, while their spatial relations determine which one is relevant. If the embedding preserves this evidence, it can separate the relevant item from distractors; otherwise, the candidates may remain difficult to distinguish. Preserving such discriminative evidence in a single vector remains a key challenge.

\begin{figure*}[t]
\centering
\includegraphics[width=\textwidth]{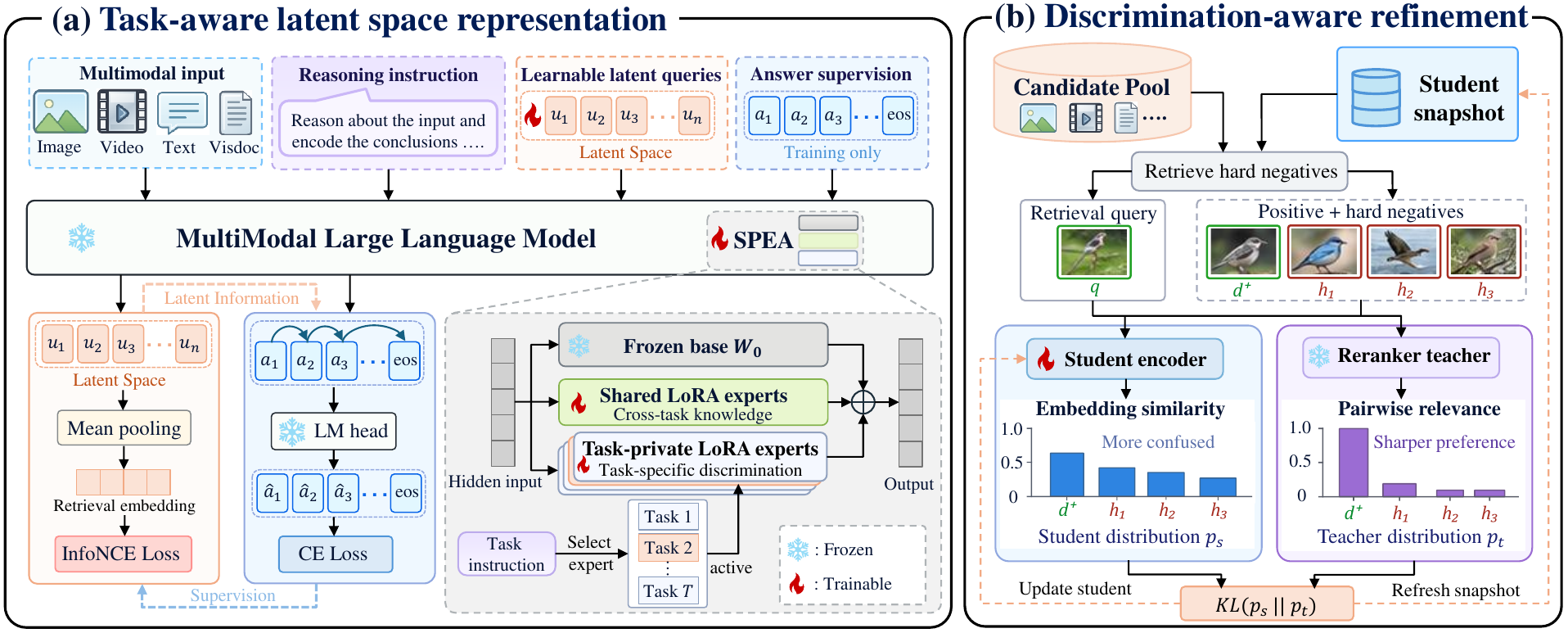}
\caption{Overview of DiLE. (a) Task-aware representation combines Task-guided Latent Space Learning and Shared--Private Expert Adaptation. (b) Discrimination-aware refinement transfers the teacher's fine-grained relevance distribution to the student through On-Policy Hard-negative Distillation. Periodically refreshed student snapshots update candidate sets, targeting evolving retrieval ambiguities.}
\label{fig:overview}
\end{figure*}

To address this challenge, we propose \method, a discrimination-aware latent-space learning framework driven by on-policy retrieval feedback. Its core idea is to use task guidance to learn \emph{what to retain} and teacher guidance on student-retrieved candidates to learn \emph{what to distinguish}. As shown in Figure~\ref{fig:overview}, \method first learns task-aware latent space representations, then performs discrimination-aware refinement through on-policy hard-negative distillation.

For task-aware latent space representation, we introduce Task-guided Latent Space Learning (TLSL). A reasoning instruction guides a small set of learnable latent queries to extract task-relevant information from each input; their hidden states are pooled into a retrieval vector. The instruction encourages the queries to focus on evidence useful for the given task. During training, auxiliary answer-generation supervision provides an additional signal that guides the latent representations toward information useful for understanding and retrieval. To accommodate task-specific discriminative needs, we also introduce Shared--Private Expert Adaptation (SPEA). Shared experts learn common knowledge across tasks, while private experts capture task-specific distinctions. This design combines knowledge sharing with the flexibility to focus on different evidence across modalities and tasks. Together, TLSL and SPEA learn task-aware latent space representations that retain relevant information and provide a foundation for subsequent discrimination-aware refinement.

For discrimination-aware refinement, we introduce On-Policy Hard-negative Distillation (OPHD), which adapts on-policy distillation~\cite{agarwal2024gkd} to retrieval. The student determines which negatives it learns from through its own retrieval policy, and the teacher provides supervision on those candidates.
A Qwen3-VL-8B reranker~\cite{qwen3-vl-reranker} jointly examines each query--candidate pair and provides a fine-grained relevance distribution that distinguishes relevant items from confusable negatives. Reverse KL aligns the student's embedding-based similarity distribution with these teacher judgments, encouraging fine-grained discrimination in the embedding space. We periodically refresh the candidates with updated student snapshots so that distillation can target the hard negatives surfaced by the evolving retrieval policy.  At inference, each input is independently encoded into a single vector, without the teacher, answer generation, or explicit reasoning traces.

We build four models using Qwen2-VL-2B/7B~\cite{wang2024qwen2} and Qwen3-VL-2B/4B~\cite{Qwen3-VL}, achieving state-of-the-art overall performance on MMEB-V2, UVRB, and MRMR. As shown in Figure~\ref{fig:teaser}(b), DiLE with Qwen3-VL-2B improves MMEB-V2 overall performance by 2.8\% over Embed-RL~\cite{embedrl} with the same backbone and is 101$\times$ faster. With Qwen2-VL-2B, DiLE improves by 4.3\% over PLUME~\cite{plume} with the same backbone and is 4$\times$ faster. These results demonstrate the effectiveness of DiLE across diverse multimodal retrieval tasks. Our contributions are:
\begin{itemize}
\item We introduce TLSL and SPEA to learn task-aware latent space representations that combine shared knowledge with task-specific distinctions.
\item We propose OPHD that aligns student similarities with teacher relevance judgments on student-retrieved hard negatives, periodically refreshing candidates to track the student's evolving retrieval behavior.
\item DiLE achieves state-of-the-art overall performance on MMEB-V2, UVRB, and MRMR, demonstrating its effectiveness across diverse modalities and retrieval tasks.
\end{itemize}

\section{Related Work}
\label{sec:related}
\noindent\textbf{Multimodal embeddings.}
Early multimodal embedding methods focus on aligning image and text inputs in a shared space~\cite{radford2021learning,jia2021scaling,zhai2023sigmoid,magiclens}. Vision--language models such as BLIP-2, LLaVA, and Qwen-VL combine visual perception with language understanding, providing a foundation for more expressive multimodal representations~\cite{li2023blip,liu2023visual,bai2023qwenvl}. Building on these capabilities, MLLM-based embedding methods support richer inputs and diverse retrieval tasks through multimodal understanding and instruction following~\cite{jiang2024e5,jiang2024vlm2vec,lin2024mm,zhang2024gme,meng2025vlm2vec,tsembed}. Further advances explore complementary aspects of retrieval learning: LamRA and LLaVE improve multimodal retrieval and hard-negative training~\cite{liu2025lamra,lan2025llave}, while MegaPairs and UniME study training data and cross-modal learning~\cite{zhou2025megapairs,gu2025breaking}. ColPali and VisRAG extend retrieval to visually rich documents~\cite{faysse2024colpali,yu2024visrag}. DiLE further combines shared and task-specific representations to capture the evidence each task requires.

\noindent\textbf{Reasoning and latent representations.}
Chain-of-thought encourages language models to generate intermediate steps for complex tasks~\cite{wei2022chain}. Recent work extends this idea to enrich multimodal embeddings. UME-R1, Think-Then-Embed, and ReTRv-R1 use generated reasoning~\cite{lan2025ume,cui2025think,zhu2025retrv}; Embed-RL aligns reasoning with retrieval~\cite{embedrl}, while RIME uses retrieval-oriented rewriting~\cite{rime}. These methods explore how intermediate information can help embeddings express the intent of a query. To reduce the cost of generating explicit reasoning, latent representations offer an alternative, as explored by Coconut and pause tokens~\cite{hao2024coconut,goyal2024pause}. For multimodal embedding, PLUME uses iterative latent states~\cite{plume}, whereas LaME and BToks supervise learnable tokens~\cite{lame,sun2026btoks}. DiLE learns single-pass latent representations using reasoning instructions and answer-generation supervision.

\noindent\textbf{Knowledge distillation.}
Knowledge distillation transfers soft teacher predictions to students~\cite{hinton2015distilling}. In retrieval, dual encoders learn from cross-encoder judgments~\cite{miech2021thinking,lei2022loopitr} through ranking scores or margins~\cite{reddi2021rankdistil,hofstatter2021margin}. CPRD emphasizes the relative ordering of informative hard negatives~\cite{chen2024cprd}, while iterative distillation uses an updated retriever to select candidates for teacher supervision~\cite{jiang-etal-2023-boot}. Rabtriever distills a generative reranker using an auxiliary token-level reverse KL loss~\cite{chen2026rabtriever}. In language modeling, on-policy distillation trains the student using teacher feedback on its own generated sequences~\cite{agarwal2024gkd}.
DiLE’s OPHD adapts on-policy distillation to retrieval by aligning student embedding similarities with the reranker’s relevance distribution via reverse KL over candidates periodically refreshed by the student’s evolving retrieval policy.

\section{Method}
\label{sec:method}
\method learns a single-vector encoder in two stages (Figure~\ref{fig:overview}). Task-guided Latent Space Learning (TLSL) and Shared--Private Expert Adaptation (SPEA) first build task-aware latent space representations. On-Policy Hard-negative Distillation (OPHD) then sharpens retrieval discrimination through teacher judgments on student-retrieved hard negatives.

\subsection{Task-Aware Latent Space Representations}
\label{sec:readout}
\noindent\textbf{TLSL.}
For a multimodal input $x$ and task instruction $i$, let $X(x,i)$ be their token sequence. We append a reasoning instruction $r$ and $N$ learnable latent queries $\mathbf U=[u_1,\ldots,u_N]$ to form $[X(x,i);r;\mathbf U]$, with $N=4$ by default. The instruction is: \emph{``Use the following latent queries to reason about the input and encode the conclusions needed to identify a matching item.''} It guides the latent queries to perform implicit reasoning over retrieval-relevant information, focusing on evidence that distinguishes matching items from confusable alternatives.
Under causal attention, each query attends to the input, instructions, and preceding queries. The latent queries act as learnable information extractors: through successive backbone layers, their hidden states repeatedly attend to input representations and progressively integrate task-relevant evidence.
We mean-pool the final hidden states of the learnable queries and normalize the result to obtain the retrieval vector:
\begin{equation}
\begin{aligned}
H &= F_\theta([X(x,i);r;\mathbf U])\\
e_\theta(x,i) &= \operatorname{norm}\!\left(
\frac{1}{N}\sum_{n=1}^{N}h_{u_n}
\right)
\end{aligned}
\label{eq:embedding}
\end{equation}
where $F_\theta$ denotes the multimodal backbone with expert adaptation, $h_{u_n}$ is the final hidden state of query $u_n$, and $\operatorname{norm}$ denotes $\ell_2$ normalization.

To guide the latent queries toward task-relevant information, we append an answer annotation $a$ after the latent queries in each training input, following Embed-RL~\cite{embedrl}. Answer tokens can attend to the input, latent queries, and preceding answer tokens. The latent queries thus provide information for answer prediction, while answer supervision backpropagates through the latent states, encouraging them to extract more accurate and useful task-relevant evidence.

We map the hidden states used for next-token prediction through the frozen language modeling head and apply cross-entropy loss only to answer tokens. We supervise both the retrieval-query side ($s=q$) and the candidate side ($s=d$):
\begin{equation}
\mathcal L_{\mathrm{ans}}^s
=
-\frac{1}{T_s}
\sum_{(x,a)\in\mathcal B_s}
\sum_{\ell=1}^{|a|}
\log P_\theta
\left(a_\ell\mid X(x,i),r,\mathbf U,a_{<\ell}\right)
\label{eq:answer}
\end{equation}
where $\mathcal B_s$ contains the input--answer pairs on side $s$ of the batch, $T_s=\sum_{(x,a)\in\mathcal B_s}|a|$ counts supervised answer tokens. The probability $P_\theta$ is obtained by applying softmax to the frozen head's output logits. Because answers follow the latent queries, causal attention prevents answer tokens from affecting the query states used for retrieval. Answer supervision is used only during training; inference encodes each input without answer tokens or answer generation.

We use InfoNCE~\cite{oord2018representation} to supervise the embedding vectors. For a batch of $B$ retrieval queries $\{q_b\}_{b=1}^{B}$, let $\mathcal D_b$ denote the in-batch candidates for $q_b$ and $\mathcal P_b\subseteq\mathcal D_b$ its labeled positives. The batch-averaged retrieval loss is
\begin{equation}
\mathcal L_{\mathrm{ret}}
=
-\frac{1}{B}\sum_{b=1}^{B}
\log
\frac{
\sum_{d\in\mathcal P_b}
\exp\!\left(\cos(\mathbf e_{q_b},\mathbf e_d)/\tau_s\right)
}{
\sum_{d\in\mathcal D_b}
\exp\!\left(\cos(\mathbf e_{q_b},\mathbf e_d)/\tau_s\right)
}
\label{eq:retrieval}
\end{equation}
where $\mathbf e_{q_b}=e_\theta(q_b,i_{q_b})$ and $\mathbf e_d=e_\theta(d,i_d)$ are the retrieval embeddings, $\cos(\cdot,\cdot)$ denotes cosine similarity, and $\tau_s$ is the temperature, set to $0.02$ by default.
Stage~I combines retrieval supervision with answer supervision on both sides:
\begin{equation}
\mathcal L_{\mathrm I}
=
\mathcal L_{\mathrm{ret}}
+
\lambda_a
\left(
\mathcal L_{\mathrm{ans}}^q
+
\mathcal L_{\mathrm{ans}}^d
\right)
\label{eq:stage1}
\end{equation}
where $\lambda_a$ controls the weight of answer supervision and is set to $0.1$ by default. The retrieval loss separates matching items from negatives, while answer supervision encourages the latent states to retain task-relevant evidence.

\noindent\textbf{SPEA.}
In UME, tasks differ in their focus and the discriminative evidence needed to identify relevant items. To accommodate these differences, we introduce SPEA, which combines shared and task-private LoRA experts~\cite{hu2022lora}. Shared experts capture cross-task knowledge, while task-private experts learn task-specific distinctions. Together, they help the latent queries extract information useful for both general task understanding and fine-grained discrimination.
The task instruction determines the private expert through a predefined mapping $t=\rho(i)$. At each adapted projection, the selected private update is combined with the shared update:
\begin{equation}
\begin{aligned}
\widetilde W_t v &= W_0v+\Delta W_s v+\Delta W_t v\\
\Delta W &= (\alpha/k)LR
\end{aligned}
\label{eq:adapt}
\end{equation}
where $W_0$ is the frozen projection, $v$ is its input, $\Delta W_s$ and $\Delta W_t$ denote the shared and selected task-private updates, and $L,R$ are rank-$k$ factors with scaling coefficient $\alpha/k$.

We define six private experts: four for image classification, image question answering, image retrieval, and image grounding, and two for video and visual documents. During training and inference, all query and candidate inputs use one shared expert with the task-private expert selected by their task instruction. We train the latent queries and adapters while keeping the backbone and language modeling head frozen.
Together, TLSL and SPEA produce task-aware latent space representations, providing the foundation for discrimination-aware refinement in Stage~II.

\subsection{Discrimination-Aware Refinement}
\label{sec:distill}
\noindent\textbf{OPHD.}
Stage~II adapts on-policy distillation to retrieval through OPHD, which distills teacher preferences over hard negatives retrieved by the student itself. We initialize the student and its first mining snapshot from the Stage~I checkpoint. For each retrieval-query $q$, the snapshot retrieves the top 20 candidates from the training corpus by cosine similarity. 
We remove known matches, related items, and duplicates. To reduce the risk of mining false negatives, we retain only high-ranked candidates whose similarity is at least $0.02$ below the mean similarity of the retrieved positives.
Up to four retained negatives are combined with the labeled positives $\mathcal P_q$ to form a distillation group $\mathcal G_q$.

Within each group, the retrieved negatives reflect the student’s current retrieval policy and may score nearly as highly as relevant items. A frozen Qwen3-VL-Reranker-8B teacher~\cite{qwen3-vl-reranker} $T$ jointly examines each query--candidate pair, accessing fine-grained interactions that can resolve these ambiguities and yield sharper relevance preferences. To transfer these preferences into the student's embedding space, we normalize student similarities and teacher relevance scores over the same group:
\begin{equation}
\begin{aligned}
p_s(d\mid q)
&=
\frac{\exp\!\left(\cos(\mathbf e_q,\mathbf e_d)/\tau_s\right)}
{\sum_{c\in\mathcal G_q}
\exp\!\left(\cos(\mathbf e_q,\mathbf e_c)/\tau_s\right)}\\
p_T(d\mid q)
&=
\frac{\exp\!\left(T(q,d)/\tau_T\right)}
{\sum_{c\in\mathcal G_q}
\exp\!\left(T(q,c)/\tau_T\right)}
\end{aligned}
\label{eq:distribution}
\end{equation}
where $\tau_s=0.02$ and $\tau_T=0.05$ are the default student and teacher temperatures. We align these distributions using reverse KL:
\begin{equation}
D_{\mathrm{KL}}(p_s\|p_T)
=
\sum_{d\in\mathcal G_q}
p_s(d\mid q)
\log\frac{p_s(d\mid q)}{p_T(d\mid q)}
\label{eq:rkl}
\end{equation}
Weighting by student probability penalizes candidates to which the student assigns substantial mass despite low teacher relevance. This encourages the student to concentrate probability on teacher-preferred candidates, transferring fine-grained discrimination into embedding similarities.

To keep distillation consistent with the labels, we apply a positive-agreement gate. Let
$d_q^\star=\operatorname*{arg\,max}_{d\in\mathcal G_q}T(q,d)$
denote the teacher's top-ranked candidate:
\begin{equation}
g_q=
\begin{cases}
1, & d_q^\star\in\mathcal P_q,\\
0, & \text{otherwise}.
\end{cases}
\label{eq:gate}
\end{equation}
Thus, a group contributes to distillation only when the teacher ranks a labeled positive first. The Stage~II objective is
\begin{equation}
\mathcal L_{\mathrm{II}}
=
\mathcal L_{\mathrm{ret}}
+
\lambda_d g_q D_{\mathrm{KL}}(p_s\|p_T)
\label{eq:stage2}
\end{equation}
where $\lambda_d=0.5$ by default. During batch aggregation, retrieval losses are averaged over all groups and distillation losses over groups passing the gate.

\begin{table*}[t]
\centering
\caption{Comparison on MMEB-V2. The top two results are highlighted with \textbf{bold} and \underline{underline}, respectively. CLS: classification, QA: question answering, RET: retrieval, GD: grounding, MRET: moment retrieval, VDR: ViDoRe, VR: VisRAG, OOD: out-of-domain.}
\label{tab:mmeb}
\renewcommand{\oursrow}{\rowcolor{gray!12}}
\scriptsize
\setlength{\tabcolsep}{2pt}
\setlength{\aboverulesep}{1.5pt}
\setlength{\belowrulesep}{1.5pt}
\renewcommand{\arraystretch}{0.98}
\def\currenttable{\begin{tabular}{ll*{16}{r}}
\toprule
\multirow{2}{*}{Model} & \multirow{2}{*}{Backbone} & \multicolumn{5}{c}{Image} & \multicolumn{5}{c}{Video} & \multicolumn{5}{c}{VisDoc} & \multirow{2}{*}{All}\\
\cmidrule(lr){3-7}\cmidrule(lr){8-12}\cmidrule(lr){13-17}
 & & CLS & QA & RET & GD & Avg. & CLS & QA & RET & MRET & Avg. & VDRv1 & VDRv2 & VR & OOD & Avg. &\\
\midrule
\multicolumn{2}{l}{\# of Datasets} & 10 & 10 & 12 & 4 & 36 & 5 & 5 & 5 & 3 & 18 & 10 & 4 & 6 & 4 & 24 & 78\\
\midrule
\multicolumn{18}{c}{\textit{$\sim$2B model size}}\\
\midrule
ColPali-V1.3~\cite{faysse2024colpali} & PaliGemma-3B & 40.3 & 11.5 & 48.1 & 40.3 & 34.9 & 26.7 & 37.8 & 21.6 & 25.5 & 28.2 & \underline{83.6} & 52.0 & 81.1 & 43.1 & 71.0 & 44.4\\
GME~\cite{zhang2024gme} & Qwen2-VL-2B & 54.4 & 29.9 & 66.9 & 55.5 & 51.9 & 34.9 & 42.0 & 25.6 & 32.4 & 33.9 & \textbf{86.1} & 54.0 & 82.5 & 43.1 & 72.7 & 54.1\\
VLM2Vec~\cite{jiang2024vlm2vec} & Qwen2-VL-2B & 58.7 & 49.3 & 65.0 & 72.9 & 59.7 & 33.4 & 30.5 & 20.6 & 33.0 & 29.0 & 49.8 & 13.5 & 51.8 & 33.5 & 41.6 & 47.0\\
VLM2Vec-V2~\cite{meng2025vlm2vec} & Qwen2-VL-2B & 62.9 & 56.3 & 69.5 & 77.3 & 64.9 & 39.3 & 34.3 & 28.8 & 38.5 & 34.9 & 75.5 & 44.9 & 79.4 & 39.4 & 65.4 & 58.0\\
UME-R1~\cite{lan2025ume} & Qwen2-VL-2B & 64.8 & 62.8 & 67.6 & 77.2 & 66.6 & 44.3 & 51.2 & 32.9 & 39.7 & 42.2 & 72.4 & 46.2 & 79.2 & 37.2 & 63.9 & 60.1\\
TTEs~\cite{cui2025think} & Qwen2-VL-2B & 67.9 & 66.6 & 70.2 & 84.1 & 70.1 & 47.3 & 49.1 & 34.4 & 33.2 & 32.1 & 77.5 & 53.2 & 83.2 & 41.1 & 68.8 & 63.1\\
RIME~\cite{rime} & Qwen2-VL-2B & 67.9 & 64.4 & 69.8 & 82.1 & 69.1 & 48.0 & 52.1 & 33.6 & 39.2 & 43.7 & 76.4 & 51.4 & 81.7 & 63.9 & 71.4 & 64.1\\
PLUME~\cite{plume} & Qwen2-VL-2B & 66.5 & 59.2 & 67.6 & 79.7 & 66.3 & 45.0 & 52.3 & 33.5 & 46.7 & 44.1 & 72.1 & 49.8 & 78.1 & 57.4 & 67.5 & 61.6\\
UMER-E~\cite{umer} & Qwen2-VL-2B & 64.4 & 64.9 & 70.2 & 78.4 & 68.0 & 43.0 & 50.6 & 34.7 & 32.8 & 41.1 & 76.1 & 50.1 & 82.8 & \underline{68.3} & 72.2 & 63.1\\
LaME~\cite{lame} & Qwen2-VL-2B & 67.6 & 66.2 & 70.5 & 81.2 & 69.3 & 45.6 & 52.3 & 36.3 & 43.5 & 44.5 & 78.2 & 48.4 & 82.8 & 55.4 & 72.1 & 64.4\\
Embed-RL~\cite{embedrl} & Qwen3-VL-2B & 62.8 & 67.9 & 68.6 & \underline{90.4} & 69.2 & \underline{57.0} & 55.9 & \underline{45.1} & \underline{49.4} & \underline{52.1} & 79.9 & 52.0 & 84.6 & 65.7 & 74.1 & 66.8\\
ReWAM~\cite{rewam} & Qwen3-VL-2B & \textbf{68.9} & \textbf{70.7} & 70.5 & 86.0 & \underline{71.9} & 53.6 & \textbf{63.3} & 38.0 & 39.9 & 49.7 & 78.6 & \textbf{55.4} & 84.2 & 66.7 & 74.2 & \underline{67.4}\\
\oursrow \textbf{\method (Ours)} & Qwen2-VL-2B & 67.7 & 66.7 & \underline{71.3} & 86.9 & 70.8 & 48.3 & 52.6 & 36.1 & 38.3 & 44.4 & 80.4 & 51.3 & \underline{85.0} & 67.8 & \underline{74.6} & 65.9\\
\oursrow \textbf{\method (Ours)} & Qwen3-VL-2B & \underline{68.8} & \underline{70.2} & \textbf{71.5} & \textbf{91.8} & \textbf{72.6} & \textbf{57.9} & \underline{60.6} & \textbf{45.8} & \textbf{50.8} & \textbf{54.1} & 81.7 & \underline{55.3} & \textbf{86.7} & \textbf{70.6} & \textbf{76.7} & \textbf{69.6}\\
\midrule
\multicolumn{18}{c}{\textit{4B--7B model size}}\\
\midrule
GME~\cite{zhang2024gme} & Qwen2-VL-7B & 57.7 & 34.7 & 71.2 & 59.3 & 56.0 & 37.4 & 50.4 & 28.4 & 38.2 & 38.6 & \textbf{89.4} & 55.6 & 85.0 & 44.4 & 75.2 & 57.8\\
LamRA-2~\cite{liu2025lamra} & Qwen2-VL-7B & 59.2 & 26.5 & 70.0 & 62.7 & 54.1 & 39.3 & 42.6 & 24.3 & 34.6 & 35.2 & 22.0 & 11.5 & 37.4 & 21.0 & 23.9 & 40.4\\
LamRA-2.5~\cite{liu2025lamra} & Qwen2.5-VL-7B & 51.7 & 34.1 & 66.9 & 56.7 & 52.4 & 32.9 & 42.6 & 23.2 & 37.6 & 33.7 & 56.3 & 33.3 & 58.2 & 40.1 & 50.2 & 47.4\\
VLM2Vec~\cite{jiang2024vlm2vec} & Qwen2-VL-7B & 62.7 & 56.9 & 69.4 & 82.2 & 65.5 & 39.1 & 30.0 & 29.0 & 40.6 & 34.0 & 56.9 & 9.4 & 59.1 & 38.1 & 46.4 & 52.3\\
VLM2Vec-V2~\cite{meng2025vlm2vec} & Qwen2-VL-7B & 65.7 & 61.5 & 70.0 & 85.2 & 68.1 & 45.9 & 33.9 & 27.6 & 39.3 & 36.4 & 78.8 & 52.6 & 82.7 & 42.1 & 69.3 & 61.2\\
CAFe~\cite{yu2025cafe} & LLaVA-OV-7B & 63.6 & 61.7 & 69.1 & 87.6 & 67.6 & 35.8 & 58.7 & 34.4 & 39.5 & 42.4 & 70.7 & 49.6 & 79.5 & 38.1 & 63.9 & 60.6\\
UME-R1~\cite{lan2025ume} & Qwen2-VL-7B & 67.1 & 69.2 & 71.9 & 84.9 & 71.3 & 48.6 & 60.7 & 38.2 & 39.3 & 47.5 & 75.7 & 50.5 & 83.7 & 37.6 & 67.1 & 64.5\\
RIME~\cite{rime} & Qwen2-VL-7B & \underline{70.3} & 71.7 & 73.2 & 86.3 & 73.4 & 52.6 & 62.0 & 38.4 & 41.6 & 49.4 & 80.9 & 55.6 & 85.8 & 66.9 & 75.6 & 68.6\\
LaME~\cite{lame} & Qwen2-VL-7B & 70.0 & 71.0 & 73.1 & 85.8 & 73.0 & 51.0 & \underline{64.8} & 40.3 & 44.6 & 50.8 & 81.5 & \underline{58.2} & 85.0 & 68.2 & \underline{75.9} & 68.8\\
Embed-RL~\cite{embedrl} & Qwen3-VL-4B & 63.7 & 70.5 & 71.3 & \underline{91.4} & 70.1 & \underline{57.6} & 58.4 & \underline{45.1} & \underline{49.5} & \underline{53.0} & 80.2 & 53.4 & 84.9 & 67.1 & 74.7 & 68.1\\
ReWAM~\cite{rewam} & Qwen3-VL-4B & \underline{70.3} & 71.8 & 72.3 & 88.4 & 73.4 & 52.2 & 63.2 & 39.5 & 40.4 & 49.8 & 80.1 & \underline{58.2} & 85.7 & 68.1 & 75.8 & 68.7\\
\oursrow \textbf{\method (Ours)} & Qwen2-VL-7B & \textbf{71.0} & \underline{72.5} & \textbf{74.6} & 90.0 & \underline{74.7} & 53.0 & 57.3 & 41.2 & 46.8 & 49.9 & 84.0 & \textbf{58.9} & \underline{86.9} & \underline{71.0} & \textbf{78.4} & \underline{70.1}\\
\oursrow \textbf{\method (Ours)} & Qwen3-VL-4B & 69.6 & \textbf{73.8} & \underline{74.1} & \textbf{93.2} & \textbf{74.9} & \textbf{59.9} & \textbf{65.9} & \textbf{48.5} & \textbf{51.7} & \textbf{57.0} & \underline{84.6} & 56.6 & \textbf{87.4} & \textbf{71.2} & \textbf{78.4} & \textbf{71.8}\\
\bottomrule
\end{tabular}}
\sbox0{\currenttable}
\setlength{\tabcolsep}{\dimexpr\tabcolsep+(\textwidth-\wd0)/36\relax}
\currenttable
\end{table*}

As training changes the student's embedding space, previously mined negatives may become easy or no longer reflect its current errors. We therefore periodically use newly captured student snapshots to re-encode the training queries and corpus and mine new hard-negative groups. This process runs asynchronously, and the frozen teacher rescores the new groups. Training continues with the available groups while the updated groups are prepared, and switches to them at fixed training milestones. By drawing new negatives from successive student snapshots, OPHD keeps teacher supervision focused on the distinctions that remain difficult under the student’s evolving retrieval policy.

\noindent\textbf{Inference.}
Each query or candidate is encoded once using one shared expert and its instruction-selected private expert, following Eq.~\ref{eq:embedding}. Corpus vectors can be precomputed, and retrieval uses dot products between embeddings, without teacher scoring, answer generation, or reasoning traces.
\label{sec:inference}

\section{Experiments}
\label{sec:experiments}
\subsection{Experimental Setup}
\label{sec:setup}
\begin{table*}[t]
\centering
\caption{Comparison on MRMR. MRMR covers 11 domains: Art, Medicine (Med.), Science (Sci.), Humanities (Hum.), Math, Physics (Phy.), Engineering (Eng.), Business (Bus.), Negation (Neg.), Design (Des.), and Traffic (Tra.).}
\label{tab:mrmr}
\renewcommand{\oursrow}{\rowcolor{gray!12}}
\scriptsize
\setlength{\tabcolsep}{3pt}
\setlength{\aboverulesep}{1.5pt}
\setlength{\belowrulesep}{1.5pt}
\renewcommand{\arraystretch}{0.98}
\def\currenttable{\begin{tabular}{llc*{12}{r}}
\toprule
\multirow{2}{*}{Model} & \multirow{2}{*}{Backbone} & \multirow{2}{*}{Size} & \multicolumn{4}{c}{Knowledge} & \multicolumn{4}{c}{Theorem} & \multicolumn{3}{c}{Contradiction} & \multirow{2}{*}{All}\\
\cmidrule(lr){4-7}\cmidrule(lr){8-11}\cmidrule(lr){12-14}
 & & & Art & Med. & Sci. & Hum. & Math & Phy. & Eng. & Bus. & Neg. & Des. & Tra. &\\
\midrule
EVA-CLIP~\cite{evaclip} & EVA-ViT & 0.4B & 10.2 & 13.5 & 26.1 & 12.9 & 6.2 & 10.5 & 9.3 & 11.7 & 8.5 & 4.4 & 5.4 & 10.8\\
OpenCLIP~\cite{cherti2023reproducible} & ViT-g/14 & 1.4B & 56.0 & 17.9 & 33.2 & 22.0 & 5.7 & 5.0 & 7.0 & 9.7 & 13.0 & 8.1 & 12.4 & 17.3\\
VISTA~\cite{vista} & BGE-visualized-m3 & 0.2B & 21.3 & 27.8 & 32.6 & 17.0 & 18.8 & 17.1 & 17.3 & 28.6 & \underline{20.0} & 20.2 & 9.4 & 20.9\\
E5-V~\cite{jiang2024e5} & LLaVA-Next & 8B & 25.1 & 11.7 & 16.6 & 10.8 & 2.1 & 3.4 & 2.5 & 5.2 & 11.5 & 3.7 & 2.1 & 8.6\\
VLM2Vec~\cite{jiang2024vlm2vec} & Qwen2-VL & 7B & 53.5 & 22.4 & 36.7 & 24.0 & 2.1 & 2.8 & 2.8 & 2.9 & 11.5 & 5.6 & 18.3 & 18.1\\
ColPali~\cite{faysse2024colpali} & PaliGemma & 3B & 36.1 & 29.9 & 42.7 & 29.2 & 5.7 & 14.8 & 12.0 & 24.6 & \textbf{28.5} & 19.4 & 18.2 & 23.7\\
GME~\cite{zhang2024gme} & Qwen2-VL & 7B & 54.3 & 40.1 & 46.8 & 45.6 & 28.8 & 36.0 & 30.2 & 45.1 & 15.0 & 26.3 & 29.6 & 36.2\\
MM-Embed~\cite{lin2024mm} & NV-Embed & 8B & 65.6 & 53.0 & 63.5 & 62.8 & 23.6 & 30.8 & 27.4 & 44.9 & 7.0 & 23.8 & 34.9 & 39.8\\
Ops-MM-Embed~\cite{opsmmembed} & Qwen2-VL & 7B & 79.3 & 52.5 & 70.0 & 67.8 & 27.7 & 39.5 & 30.1 & 52.3 & 8.0 & 55.9 & \underline{45.8} & 48.1\\
Embed-RL~\cite{embedrl} & Qwen3-VL & 4B & 73.5 & 53.1 & 60.5 & 69.8 & 32.9 & 45.6 & 35.1 & 52.6 & 6.0 & 54.3 & 31.7 & 46.8\\
UME-R1~\cite{lan2025ume} & Qwen2-VL & 7B & 77.8 & 55.7 & 72.9 & 64.1 & 27.2 & 39.2 & 32.2 & 47.8 & 7.5 & 61.9 & 41.7 & 48.0\\
LaME~\cite{lame} & Qwen2-VL & 7B & 73.4 & 58.2 & 73.8 & 65.6 & 29.5 & 44.4 & 36.4 & 52.2 & 8.5 & \underline{64.9} & 40.9 & 49.8\\
RIME~\cite{rime} & Qwen2-VL & 7B & 76.8 & 58.5 & 73.6 & 71.4 & 29.2 & 43.0 & 35.6 & 51.8 & 8.5 & 64.1 & 39.5 & 50.2\\
ReWAM~\cite{rewam} & Qwen3-VL & 4B & 79.9 & 60.0 & 73.0 & 74.8 & \underline{40.2} & \underline{52.2} & \textbf{43.1} & \textbf{56.0} & 6.5 & 46.7 & \textbf{53.5} & \underline{53.3}\\
\midrule
\oursrow \textbf{\method (Ours)} & Qwen2-VL & 2B & 76.0 & 56.9 & 72.1 & 66.6 & 27.8 & 38.5 & 31.8 & 50.5 & 11.0 & 44.3 & 29.7 & 45.9\\
\oursrow \textbf{\method (Ours)} & Qwen2-VL & 7B & \underline{80.2} & \underline{64.6} & \underline{78.1} & \underline{76.7} & 35.2 & 44.7 & 34.8 & \underline{54.0} & 10.5 & \textbf{65.6} & 37.6 & 52.9\\
\oursrow \textbf{\method (Ours)} & Qwen3-VL & 2B & 79.0 & 62.1 & 74.6 & 70.8 & 33.5 & 44.1 & 36.3 & 53.7 & 11.0 & 53.6 & 27.0 & 49.6\\
\oursrow \textbf{\method (Ours)} & Qwen3-VL & 4B & \textbf{82.1} & \textbf{68.3} & \textbf{80.1} & \textbf{80.1} & \textbf{44.1} & \textbf{53.2} & \underline{39.2} & \textbf{56.0} & 7.0 & 64.6 & 43.9 & \textbf{56.2}\\
\bottomrule
\end{tabular}}
\sbox0{\currenttable}
\setlength{\tabcolsep}{\dimexpr\tabcolsep+(\textwidth-\wd0)/30\relax}
\currenttable
\end{table*}
\begin{table*}[t]
\centering
\definecolor{tablebest}{RGB}{180,55,55}
\definecolor{tablethird}{RGB}{55,135,75}
\caption{Comparison on UVRB across 16 datasets, spanning three tasks: Textual (Txt.), Composed (Cmp.), and Visual (Vis.); three domains: Coarse-Grained (Coarse-G.), Fine-Grained (Fine-G.), and Long-Context (Long-Ctx.); and three sub-domains: Spatial (Spa.), Temporal (Temp.), and Partially Relevant (PR.). Avg. is the mean across all 16 datasets.}
\label{tab:uvrb}
\renewcommand{\oursrow}{\rowcolor{gray!12}}
\scriptsize
\setlength{\tabcolsep}{2pt}
\setlength{\aboverulesep}{1.5pt}
\setlength{\belowrulesep}{1.5pt}
\renewcommand{\arraystretch}{0.98}
\def\currenttable{\begin{tabular}{llc*{10}{r}}
\toprule
\multirow{2}{*}{Model} & \multirow{2}{*}{Backbone} & \multirow{2}{*}{Size} & \multicolumn{3}{c}{Tasks} & \multicolumn{3}{c}{Domains} & \multicolumn{3}{c}{Sub-domains} & \multirow{2}{*}{Avg.}\\
\cmidrule(lr){4-6}\cmidrule(lr){7-9}\cmidrule(lr){10-12}
 & & & Txt. & Cmp. & Vis. & Coarse-G. & Fine-G. & Long-Ctx. & Spa. & Temp. & PR. &\\
\midrule
InternVideo2-6B~\cite{wang2024internvideo2} & InternVideo2 & 6B & 44.8 & 22.0 & 66.0 & 50.4 & 41.7 & 42.3 & 63.1 & 40.0 & 22.0 & 42.7\\
GME~\cite{zhang2024gme} & Qwen2-VL & 2B & 53.9 & \textbf{34.5} & 59.7 & 46.1 & 47.1 & 68.5 & 71.6 & 34.9 & 34.7 & 48.8\\
UNITE~\cite{kong2025modality} & Qwen2-VL & 2B & 53.6 & 24.2 & 65.4 & 45.5 & 47.1 & 68.1 & 72.5 & 34.7 & 34.1 & 48.0\\
VLM2Vec-V2~\cite{meng2025vlm2vec} & Qwen2-VL & 2B & 58.7 & 26.3 & 61.3 & 49.8 & 50.2 & 76.2 & 80.9 & 34.8 & 34.8 & 50.8\\
BGE-VL~\cite{zhou2025megapairs} & LLaVA-1.6 & 7B & 49.7 & 26.8 & 62.2 & 44.8 & 40.6 & 63.6 & 66.4 & 29.2 & 26.1 & 44.3\\
UniME~\cite{gu2025breaking} & LLaVA-OV & 7B & 56.1 & 30.8 & 70.2 & 50.0 & 51.8 & 66.4 & 78.5 & 39.6 & 37.3 & 52.1\\
B3~\cite{thirukovalluru2025breaking} & Qwen2-VL & 7B & 57.0 & 27.0 & 67.8 & 48.2 & 50.5 & 72.2 & 79.7 & 36.4 & 35.5 & 51.1\\
GME~\cite{zhang2024gme} & Qwen2-VL & 7B & 60.4 & \underline{34.1} & 61.5 & 51.8 & 50.7 & 78.8 & 74.9 & 37.3 & 39.8 & 53.0\\
UNITE~\cite{kong2025modality} & Qwen2-VL & 7B & 60.9 & 25.4 & 66.6 & 54.1 & 53.9 & 74.6 & 77.9 & 41.2 & 42.5 & 53.8\\
UME-R1~\cite{lan2025ume} & Qwen2-VL & 7B & 59.1 & 15.9 & 66.4 & 55.5 & 53.9 & 82.5 & 82.5 & 42.7 & 41.8 & 54.6\\
RIME~\cite{rime} & Qwen2-VL & 7B & 60.8 & 22.0 & 67.5 & 56.2 & 54.0 & 81.6 & 79.9 & 43.6 & 41.3 & 55.6\\
Embed-RL~\cite{embedrl} & Qwen3-VL & 2B & 61.1 & 20.1 & 70.3 & 59.1 & 54.6 & 86.9 & 87.2 & 43.0 & 40.6 & 58.7\\
Embed-RL~\cite{embedrl} & Qwen3-VL & 4B & 62.0 & 18.4 & 70.3 & 60.7 & 55.6 & 86.1 & 87.9 & 46.0 & 40.6 & \underline{60.2}\\
\midrule
\oursrow \textbf{\method (Ours)} & Qwen2-VL & 2B & 59.3 & 16.1 & 66.5 & 53.4 & 53.8 & 70.9 & 80.8 & 39.9 & 40.6 & 51.9\\
\oursrow \textbf{\method (Ours)} & Qwen2-VL & 7B & 64.3 & 18.6 & 71.4 & 58.6 & 58.6 & 75.6 & 84.3 & 46.0 & \underline{45.6} & 56.5\\
\oursrow \textbf{\method (Ours)} & Qwen3-VL & 2B & \underline{69.9} & 23.5 & \underline{73.7} & \underline{61.9} & \underline{60.3} & \underline{87.5} & \underline{88.7} & \underline{47.5} & 44.8 & 60.1\\
\oursrow \textbf{\method (Ours)} & Qwen3-VL & 4B & \textbf{72.0} & 24.0 & \textbf{75.7} & \textbf{64.5} & \textbf{62.8} & \textbf{88.8} & \textbf{90.1} & \textbf{50.7} & \textbf{47.7} & \textbf{62.2}\\
\bottomrule
\end{tabular}}
\sbox0{\currenttable}
\setlength{\tabcolsep}{\dimexpr\tabcolsep+(\textwidth-\wd0)/26\relax}
\currenttable
\vspace{-3mm}
\end{table*}
\noindent\textbf{Implementation.}
We train four DiLE variants based on Qwen3-VL-2B/4B and Qwen2-VL-2B/7B~\cite{Qwen3-VL,wang2024qwen2}. Both training stages use AdamW with a learning rate of $5\times10^{-5}$, linear decay, 100 warm-up steps, and weight decay of $0.01$.
Stage~I trains on 1.4M samples for 5,000 steps with a global batch size of 1,024.
Stage~II uses the resulting checkpoint to initialize the student and mine the initial hard negatives. It trains on 300k queries in three successive 100k-query blocks, with a global batch size of 128 query groups. Qwen3-VL-Reranker-8B~\cite{qwen3-vl-reranker} serves as the frozen teacher. We trigger asynchronous mining after 50k and 150k processed queries, with 100k queries between refreshes. Mining runs alongside training, and training switches to refreshed candidate groups at the 100k and 200k boundaries.

\noindent\textbf{Datasets and comparisons.}
We evaluate DiLE on three complementary benchmarks. MMEB-V2~\cite{meng2025vlm2vec} measures universal multimodal retrieval across 78 tasks: 36 image, 18 video, and 24 visual-document tasks.
These cover classification, question answering, retrieval, grounding, and document understanding. 
We report Hit@1 for image and video tasks, linear-gain nDCG@5 for visual-document tasks, and the unweighted mean across all 78 tasks.
MRMR~\cite{mrmr} evaluates retrieval across 11 tasks involving knowledge, theorem application, and contradiction recognition. We report nDCG@10 for all tasks except Negation, which uses Hit@1, together with the overall benchmark score. 
UVRB~\cite{guo2025towards} evaluates video retrieval on 16 datasets with textual, composed, and visual queries. It covers coarse-grained, fine-grained, and long-context retrieval. We use the recall or precision metric specified for each dataset and report the unweighted 16-dataset mean. 
Our training data follows VLM2Vec-V2~\cite{meng2025vlm2vec}: image-based vision--language data from MMEB-train, video-language instruction data from LLaVA-Hound~\cite{llava-hound}, and visual-document retrieval data from ViDoRe~\cite{faysse2024colpali} and VisRAG~\cite{yu2024visrag}. Stage~I adds answer annotations from Embed-RL~\cite{embedrl} to supervise the latent representations. Stage~II mines hard negatives from the same training sources. 

We compare with a broad range of methods, including VLM2Vec-V2, UME-R1, PLUME, and Embed-RL~\cite{meng2025vlm2vec,lan2025ume,plume,embedrl}, among others listed in Tables~\ref{tab:mmeb}, \ref{tab:mrmr}, and~\ref{tab:uvrb}. These methods cover direct embedding learning, explicit and latent reasoning, and reinforcement-learning-based approaches. We focus on configurations using the same training sources as VLM2Vec-V2 without additional datasets. 

\subsection{Main Results}
\label{sec:results}
\noindent\textbf{MMEB-V2.}
As shown in Table~\ref{tab:mmeb}, \method achieves state-of-the-art overall performance in both model-size groups, scoring 69.6\% with Qwen3-VL-2B and 71.8\% with Qwen3-VL-4B. With matched Qwen3-VL backbones, \method outperforms the explicit-reasoning methods ReWAM and Embed-RL by 2.2\% and 2.8\% at 2B, and by 3.1\% and 3.7\% at 4B, respectively. The 4B variant also leads all three modality averages, exceeding Embed-RL by 4.8\% on images, 4.0\% on videos, and 3.7\% on visual documents. These gains show that compact latent representations can support strong retrieval discrimination without generating explicit reasoning traces.
With Qwen2-VL-2B, \method surpasses the latent-reasoning methods PLUME and LaME by 4.3\% and 1.5\% overall, respectively. 
With Qwen2-VL-7B, \method exceeds LaME by 1.3\% overall, gaining 1.7\% on images and 2.5\% on visual documents. Together, these results demonstrate the effectiveness of DiLE across diverse retrieval tasks, modalities, model scales, and backbone generations.

\noindent\textbf{MRMR.}
As shown in Table~\ref{tab:mrmr}, \method achieves state-of-the-art overall performance on reasoning-intensive retrieval. Our Qwen2-VL-2B/7B and Qwen3-VL-2B/4B variants achieve overall scores of 45.9\%, 52.9\%, 49.6\%, and 56.2\%, respectively. With Qwen2-VL-7B, \method outperforms RIME and UME-R1 with the same backbone by 2.7\% and 4.9\%. With Qwen3-VL-4B, it exceeds Embed-RL and ReWAM by 9.4\% and 2.9\%, respectively. The 4B variant leads all four Knowledge subtasks, ranks first in Math, Physics and Business. Our Qwen2-VL-7B variant also achieves the best Design score of 65.6\%. Without benchmark-specific fine-tuning, these results support the effectiveness of our task-guided latent representations on reasoning-intensive retrieval.

\begin{table}[t]
\centering
\caption{Ablation studies on task-aware representation. $\Delta$ denotes the average performance change compared with the baseline.}
\label{tab:ablations}
\begingroup
\scriptsize
\setlength{\tabcolsep}{3pt}
\setlength{\aboverulesep}{1.5pt}
\setlength{\belowrulesep}{1.5pt}
\renewcommand{\arraystretch}{0.98}
\def\currenttable{%
\begin{tabular}{clrrrrr}
\toprule
\# & Method & Image & Video & Visdoc & All & $\boldsymbol{\Delta}$\\
\midrule
\rowcolor{gray!12} 1 & Baseline  & 71.4 & 52.4 & 75.0 & 68.1 & -\\
\midrule
\multicolumn{7}{c}{\textit{TLSL structure ablation}}\\
2 & W/o R-ins  & 71.0 & 51.9 & 74.9 & 67.8 & \textbf{-0.3}\\
3 & W/o R-ins \& A-sup  & 70.6 & 51.6 & 74.1 & 67.3 & \textbf{-0.8}\\
4 & W Last-token  & 70.3 & 51.7 & 73.5 & 67.0 & \textbf{-1.1}\\
\midrule
\multicolumn{7}{c}{\textit{Number of latent queries ablation}}\\
5 & 2 queries  & 70.8 & 51.8 & 75.0 & 67.7 & \textbf{-0.4}\\
6 & 8 queries  & 71.1 & 52.2 & 74.9 & 67.9 & \textbf{-0.2}\\
\midrule
\multicolumn{7}{c}{\textit{SPEA ablation}}\\
7 & Only shared  & 70.5 & 50.5 & 75.1 & 67.3 & \textbf{-0.8}\\
8 & Only private  & 70.7 & 46.1 & 74.4 & 66.1 & \textbf{-2.0}\\
9 & 3 private  & 69.9 & 52.2 & 75.0 & 67.4 & \textbf{-0.7}\\
\bottomrule
\end{tabular}%
}
\sbox0{\currenttable}
\setlength{\tabcolsep}{\dimexpr\tabcolsep+(\columnwidth-\wd0)/14\relax}
\currenttable
\endgroup
\end{table}

\begin{table}[t]
\centering
\caption{Component ablations of OPHD. }
\label{tab:distillation_components}
\begingroup
\scriptsize
\setlength{\tabcolsep}{2pt}
\setlength{\aboverulesep}{1.5pt}
\setlength{\belowrulesep}{1.5pt}
\renewcommand{\arraystretch}{0.98}
\def\currenttable{\begin{tabular}{cccrrrrr}
\toprule
Refresh & Teacher & Gate & Image & Video & Visdoc & All & $\boldsymbol{\Delta}$\\
\midrule
\rowcolor{gray!12} $\checkmark$ & $\checkmark$ & $\checkmark$ & 72.6 & 54.1 & 76.7 & 69.6 & -\\
$\checkmark$ & $\checkmark$ & $\times$ & 72.3 & 53.9 & 76.8 & 69.4 & \textbf{-0.2}\\
$\checkmark$ & $\times$ & $\times$ & 71.8 & 53.4 & 75.9 & 68.8 & \textbf{-0.8}\\
$\times$ & $\checkmark$ & $\times$ & 72.1 & 53.3 & 76.5 & 69.1 & \textbf{-0.5}\\
$\times$ & $\times$ & $\times$ & 71.8 & 52.2 & 75.8 & 68.5 & \textbf{-1.1}\\
\bottomrule
\end{tabular}}
\sbox0{\currenttable}
\setlength{\tabcolsep}{\dimexpr\tabcolsep+(\columnwidth-\wd0)/16\relax}
\currenttable
\endgroup
\vspace{-2mm}
\end{table}

\noindent\textbf{UVRB.}
As shown in Table~\ref{tab:uvrb}, \method achieves state-of-the-art overall performance on UVRB, with a 16-dataset mean of 62.2\%. With Qwen2-VL-2B, it scores 51.9\%, exceeding VLM2Vec-V2 by 1.1\%. With Qwen2-VL-7B, it reaches 56.5\%, outperforming UME-R1 and RIME by 1.9\% and 0.9\%, respectively. With Qwen3-VL-2B and Qwen3-VL-4B, \method achieves 60.1\% and 62.2\%, surpassing Embed-RL with the same backbones by 1.4\% and 2.0\%. Both Qwen3-VL variants improve over Embed-RL across all reported task, domain, and subdomain averages. The 4B variant leads textual and visual retrieval, all three domain averages, and all three fine-grained subdomains. Its fine-grained score exceeds Embed-RL-4B by 7.2\%, with gains of 4.7\% and 7.1\% in temporal and partially relevant retrieval. These results support the effectiveness of discrimination-aware latent-space learning in capturing distinctions needed for diverse video retrieval tasks.

\subsection{Ablation Studies}
\label{sec:ablation}
We evaluate the contributions of DiLE's components and its generalization. For the TLSL and SPEA ablations in Table~\ref{tab:ablations}, the baseline is Qwen3-VL-2B trained with Stage~I alone. For the OPHD ablations in Tables~\ref{tab:distillation_components} and~\ref{tab:distillation_settings}, the baseline is the full two-stage DiLE model with the same backbone. We evaluate both baselines on MMEB-V2. Table~\ref{tab:generalization} further examines DiLE's ability to generalize.

\noindent\textbf{TLSL.}
We ablate the structure of TLSL and the number of latent queries in Table~\ref{tab:ablations}, rows~\#2--\#6. Removing the reasoning instruction (\#2) reduces the overall score by 0.3\%. Further removing answer supervision (\#3) increases the drop to 0.8\%. Both variants perform worse than the baseline across all three modalities, supporting the contributions of reasoning guidance and answer supervision to task-aware representation learning. Replacing latent-query pooling with the last-token representation (\#4) reduces the overall score by 1.1\%, highlighting the benefit of learnable queries for extracting retrieval-relevant information. Reducing the number of queries from four to two (\#5) lowers the score by 0.4\%, while increasing it to eight (\#6) yields a 0.2\% drop. Four queries thus provide the best performance among the tested settings. Together, these results support the effectiveness of TLSL in capturing task-relevant evidence within a compact latent space.

\noindent\textbf{SPEA.}
UME tasks require different evidence to identify relevant items. SPEA addresses this variation by combining shared LoRA experts that capture cross-task knowledge with private LoRA experts that learn task-specific distinctions. In Table~\ref{tab:ablations}, we vary the expert configuration in rows~\#7--\#9 while keeping the total number of trainable parameters equal to that of the baseline (\#1). Using only shared experts (\#7) lowers the overall score by 0.8\%, while using only private experts (\#8) lowers it by 2.0\%. In row~\#9, we replace the six private experts with three, one each for images, videos, and visual documents. Merging the four image-task experts into one lowers the image score from 71.4\% to 69.9\% and the overall score by 0.7\%. At the same parameter budget, these results favor combining shared and private experts and retaining separate experts for individual image tasks.

\begin{table}[t]
\centering
\caption{Ablations of the refresh schedule and distillation objective.}
\label{tab:distillation_settings}
\begingroup
\scriptsize
\setlength{\tabcolsep}{2pt}
\setlength{\aboverulesep}{1.5pt}
\setlength{\belowrulesep}{1.5pt}
\renewcommand{\arraystretch}{0.98}
\def\currenttable{\begin{tabular}{clrrrrr}
\toprule
\# & Method & Image & Video & Visdoc & All & $\boldsymbol{\Delta}$\\
\midrule
\rowcolor{gray!12} 1 & Baseline & 72.6 & 54.1 & 76.7 & 69.6 & -\\
\midrule
\multicolumn{7}{c}{\textit{Refresh schedule ablation}}\\
2 & 50k queries & 72.5 & 54.8 & 76.7 & 69.7 & \textbf{+0.1}\\
3 & 150k queries & 72.7 & 53.3 & 76.5 & 69.4 & \textbf{-0.2}\\
\midrule
\multicolumn{7}{c}{\textit{Distillation objective ablation}}\\
4 & FKL & 72.2 & 54.3 & 76.5 & 69.4 & \textbf{-0.2}\\
5 & JSD & 72.0 & 53.8 & 76.6 & 69.2 & \textbf{-0.4}\\
\bottomrule
\end{tabular}}
\sbox0{\currenttable}
\setlength{\tabcolsep}{\dimexpr\tabcolsep+(\columnwidth-\wd0)/14\relax}
\currenttable
\endgroup
\end{table}

\begin{table}[t]
\centering
\caption{Generalization of DiLE to EmbeddingGemma2. }
\label{tab:generalization}
\begingroup
\scriptsize
\setlength{\tabcolsep}{2pt}
\setlength{\aboverulesep}{1.5pt}
\setlength{\belowrulesep}{1.5pt}
\renewcommand{\arraystretch}{0.98}
\def\currenttable{\begin{tabular}{lcrrrrr}
\toprule
Method & Parameters & Image & Video & Visdoc & All & $\boldsymbol{\Delta}$\\
\midrule
EmbeddingGemma2~\cite{embeddinggemma2} & 438M & 57.3 & 50.7 & 67.8 & 59.0 & -\\
\midrule
\rowcolor{gray!12} EmbeddingGemma2\textsubscript{+DiLE} & 474M & 66.1 & 44.2 & 70.5 & 62.4 & \textbf{+3.4}\\
\bottomrule
\end{tabular}}
\sbox0{\currenttable}
\setlength{\tabcolsep}{\dimexpr\tabcolsep+(\columnwidth-\wd0)/14\relax}
\currenttable
\endgroup
\end{table}

\noindent\textbf{OPHD.}
Table~\ref{tab:distillation_components} examines three components of OPHD in Stage~II. \emph{Refresh} periodically updates hard negatives using student snapshots. \emph{Teacher} distills relevance scores from the frozen reranker. \emph{Gate} applies distillation only when the teacher ranks a labeled positive first. Full OPHD achieves an overall score of 69.6\% (row~1). Removing the gate lowers the score by 0.2\% (row~2). Starting from this ungated setting, removing the teacher yields 68.8\% (row~3), while using fixed hard negatives with the teacher yields 69.1\% (row~4). These scores are 0.8\% and 0.5\% below full OPHD, respectively. Training with fixed hard negatives alone scores 68.5\% (row~5), a drop of 1.1\%. Comparing rows~2--5 shows how refresh and distillation complement each other. Refresh keeps supervision focused on candidates the evolving student finds confusable, while the teacher provides fine-grained relevance judgments over those candidates.

Table~\ref{tab:distillation_settings} examines the refresh interval and distillation objective. Shortening the default interval from 100k to 50k queries improves the overall score by only 0.1\% (\#2). Extending it to 150k queries lowers the score by 0.2\% (\#3). We therefore use 100k queries as the default interval. For distillation, reverse KL (RKL) weights candidates by student probability, penalizing hard negatives that the student scores highly despite low teacher relevance. Forward KL (FKL) weights candidates by teacher probability, encouraging the student to cover the probability mass assigned by the teacher. JSD uses the midpoint $m=(p_s+p_T)/2$ and equally weights $D_{\mathrm{KL}}(p_s\|m)$ and $D_{\mathrm{KL}}(p_T\|m)$. Replacing RKL with FKL (\#4) or JSD (\#5) lowers the overall score by 0.2\% or 0.4\%, respectively. RKL performs best among the tested objectives, consistent with its emphasis on correcting the student's own retrieval confusions.

\begin{figure}[t]
\centering
\includegraphics[width=\linewidth]{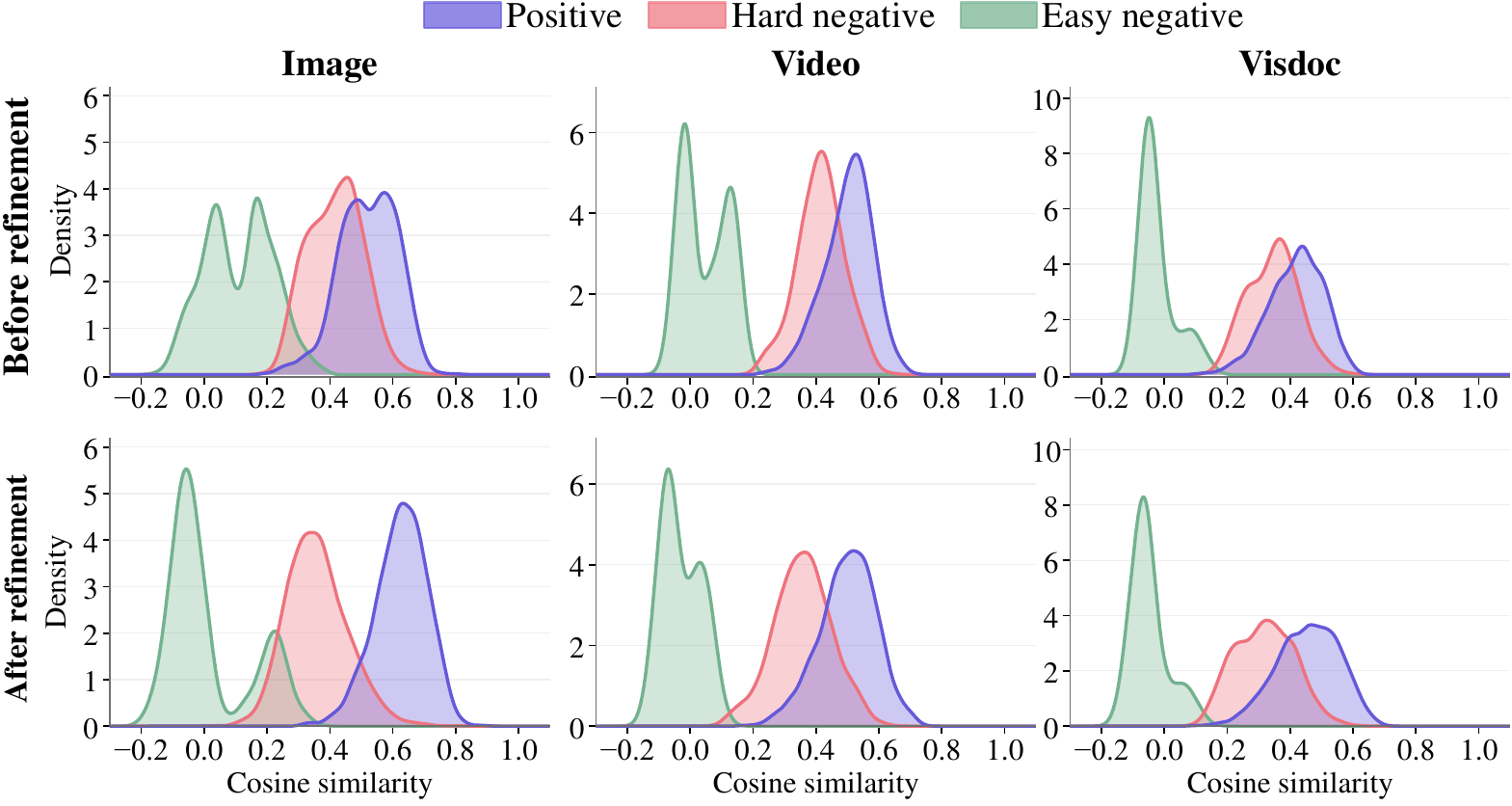}
\caption{Positive--negative similarity distributions before and after Stage~II with Qwen3-VL-2B. Refinement further separates the positive and negative distributions across three tasks.}
\label{fig:discrimination}
\vspace{-3mm}
\end{figure}

\noindent\textbf{Generalization to EmbeddingGemma2}
\label{sec:generalization}
To test DiLE's ability to generalize, we apply TLSL, SPEA, and OPHD to EmbeddingGemma2~\cite{embeddinggemma2}. As shown in Table~\ref{tab:generalization}, DiLE adds 36M parameters and improves the overall MMEB-V2 score by 3.4\%, from 59.0\% to 62.4\%. The largest gain is on image tasks, where the score rises from 57.3\% to 66.1\%. Visual-document retrieval also improves from 67.8\% to 70.5\%, while the video score drops from 50.7\% to 44.2\%. The overall gain shows that DiLE's combination of task-guided representation learning and discrimination-aware refinement also benefits a compact embedding model.

\subsection{Analyzing Discrimination and Refresh}
\label{sec:discrimination}
\begin{figure}[t]
\centering
\includegraphics[width=\linewidth]{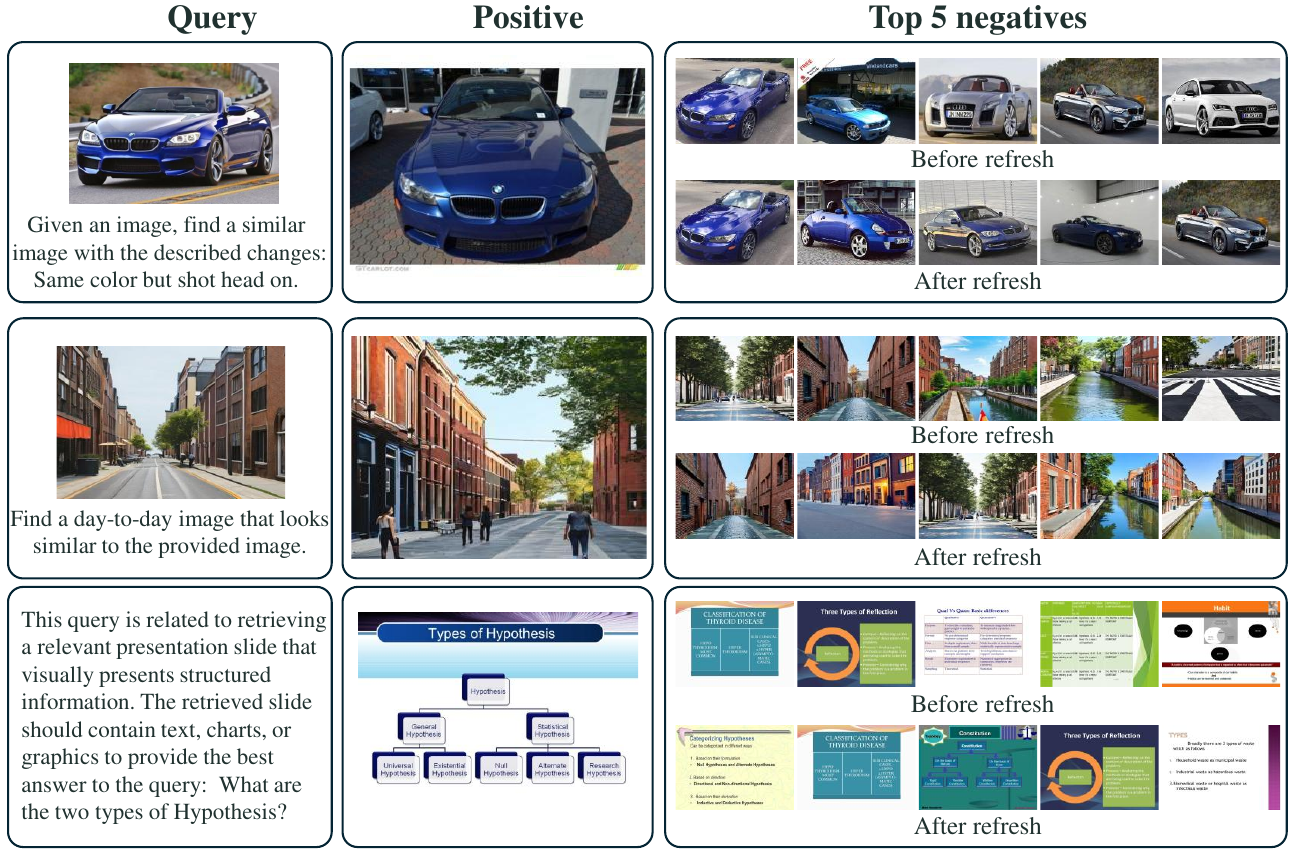}
\caption{Qualitative examples of hard-negative refresh. The updated student retrieves new confusable candidates, keeping supervision focused on the distinctions it currently struggles to capture.}
\label{fig:qualitative}
\vspace{-3mm}
\end{figure}

Figure~\ref{fig:discrimination} compares cosine similarity distributions before and after Stage~II refinement using Qwen3-VL-2B. We sample 5,000 queries from each of the image, video, and visual-document tasks. For each query, we compute its cosine similarity to positive and negative candidates. Among the negatives, we select the five with the highest similarities as hard negatives and the five with the lowest similarities as easy negatives. Across all three modalities, the positive and hard-negative distributions overlap less after refinement. The separation is most pronounced on image tasks, where the two distributions show distinct peaks after Stage~II. 
These results show that OPHD improves the model’s ability to distinguish positives from hard negatives in the embedding space.
Figure~\ref{fig:qualitative} shows how refresh changes the retrieved negatives. In the car example, refreshed candidates concentrate more on similarly colored vehicles, making viewpoint an important distinction. By repeatedly selecting candidates in the updated embedding space, refresh directs teacher feedback toward the student's current confusions, complementing distillation as shown in Table~\ref{tab:distillation_components}.

\section{Conclusion}
We present DiLE, a two-stage framework for universal multimodal embeddings. Stage~I combines task-guided latent space learning with shared--private expert adaptation to encode retrieval evidence in compact embeddings that capture both cross-task knowledge and task-specific distinctions. Stage~II refines these embeddings through on-policy distillation of teacher relevance judgments on student-retrieved hard negatives. DiLE achieves state-of-the-art overall performance on MMEB-V2, MRMR, and UVRB, demonstrating its effectiveness across diverse multimodal retrieval tasks.

\label{maintextend}
\clearpage
{\small
\bibliographystyle{ieeenat_fullname}
\bibliography{main}
}
\end{document}